\documentclass[preprint,12pt]{elsarticle}
\usepackage{amssymb}
\usepackage{amsmath}
\usepackage{comment}
\usepackage{url}
\usepackage{hyperref}
\usepackage[nomarkers,figuresfirst,nofiglist,notablist]{endfloat}

\usepackage{tikz}
\usepackage[dvipsnames]{xcolor}
\usepackage{multirow}
\usepackage{ulem}
\renewcommand{\paragraph}[1]{}
\usepackage{comment}

\newtheorem{definition}{Definition}

\begin{document}

\begin{frontmatter}



\title{DeXposure-FM: A Time-series, Graph Foundation Model for Credit Exposures and Stability on Decentralised Financial Networks}

\author[label1]{Aijie Shu}%\ead{v1ashu@ed.ac.uk}
\author[label2]{Wenbin Wu}
%\author[label2]{Bryan Zhang}
\author[label4]{Gbenga Ibikunle}%\ead{gbenga.Ibikunle@ed.ac.uk}
\author[label1]{Fengxiang He\corref{cor1}}\ead{F.He@ed.ac.uk}
\cortext[cor1]{Corresponding author}

\affiliation[label1]{organization={School of Informatics, University of Edinburgh},
%%             city={Edinburgh},
             country={United Kingdom}}

\affiliation[label2]{organization={Cambridge Centre for Alternative Finance, Judge Business School, University of Cambridge},
             country={United Kingdom}}

\affiliation[label4]{organization={Business School, University of Edinburgh},
             country={United Kingdom}}

%% Abstract
\begin{abstract}
Credit exposure in Decentralized Finance (DeFi) is implicit and token-mediated, so shocks to widely held tokens can cascade across protocols and chains. We introduce DeXposure-FM, a domain-specific time-series graph foundation model (GFM) for forecasting inter-protocol credit exposure. A GraphPFN graph-tabular encoder is fine-tuned end-to-end on 43.7 million DeXposure observations covering 4,300+ protocols, 602 chains, and 24,300+ tokens, and exposes the learned representation through lightweight multi-horizon forecasting heads whose predicted graphs feed a shared downstream measurement layer. On multi-step forecasting and predictive contagion stress testing, DeXposure-FM improves edge-existence prediction and reduces magnitude RMSE over neural baselines, while persistence remains competitive on stable magnitudes. Forecasted graphs further support macroprudential indicators, including systemic-importance scores, sector spillovers, concentration, and scenario stress losses. Gains concentrate in structural-change and tail-error regimes where carry-forward assumptions are least reliable.

\noindent\textit{Model:} \url{https://huggingface.co/EVIEHub/DeXposure-FM}.

\noindent\textit{Code:} \url{https://github.com/EVIEHub/graph-dexposure}.

\end{abstract}


\begin{highlights}
\item Domain-specific graph foundation model: We formulate DeFi credit exposures as dynamic graph-tabular financial networks and build DeXposure-FM around a GraphPFN-based encoder.

\item Empirical verification on machine learning benchmarks: On the strict 2025 hold-out, DeXposure-FM achieves AUROC 0.993--0.995 and AUPRC 0.967--0.973 for edge existence, and reduces magnitude RMSE over neural baselines while persistence remains competitive on stable magnitudes.

\item Financial economics tools: Forecasted graphs feed deterministic measurements of protocol-level systemic importance, sector-to-sector spillover concentration, and DebtRank-style scenario losses; empirical gains concentrate in worst-20\% persistence-error regimes and bridge-sector shocks.

\item Open-source efforts and reproducibility: Model checkpoints and code are publicly available at the URLs reported in the abstract.

%\item Economic measurement and systemic-risk statistics for DeFi: We use DeXposure-FM to derive richer systemic-risk measures: protocol-level systemic-importance scores, sectoral spillover indices, and early-warning indicators based on shifts in network structure and reliance on fragile collateral-providing information beyond standard size- and TVL-based metrics and supporting macroprudential surveillance and DeFi regulation.

%\item Financial risk stress testing: Using the model, we generate shock-consistent scenarios for major DeFi stress events (e.g. stablecoin de-pegs, exchange failures) and quantify loss distributions, contagion paths, and the impact of counterfactual policy tools (leverage caps, collateral haircuts, bridge limits).
\end{highlights}

%% Keywords

\end{frontmatter}


\newpage
\tableofcontents

\section{Introduction}

% Paragraph 1: Introduce DeFi networks, credit exposures, risks, and measurement gaps

Decentralized finance (DeFi) has emerged as an important and fast-evolving financial ecosystem, offering lending, trading, derivatives, and payment-like services via smart contracts deployed across multiple blockchains \cite{Schar2021DeFi,gogol2024sokdecentralizedfinancedefi}. A defining feature of DeFi is composability: protocols routinely hold, lock, or accept tokens issued by other protocols as collateral, liquidity, or reserves. As a result, credit exposure is often implicit and token-mediated rather than expressed as bilateral contracts, creating a dynamic network of inter-protocol dependencies \cite{wu2025dexposuredatasetbenchmarksinterprotocol,BERTOMEU2024105321}. When a token that is widely used as collateral or liquidity experiences a shock, through a price collapse, governance failure, exploit, or liquidation event, losses can propagate to protocols that rely on that token, potentially generating amplification and contagion across chains \cite{gogol2024sokdecentralizedfinancedefi,aufiero2025mappingmicroscopicsystemicrisks}. 
Policy authorities have highlighted the vulnerabilities of DeFi networks %In particular, collateralised lending protocols rely on oracles and liquidation mechanisms that can trigger rapid deleveraging and fire-sale dynamics when prices move sharply or data feeds are disrupted \cite{DuleyEtAl2023Oracle,SasiBrodeskyNassr2023DeFiLiquidations,TianZhu2025Liquidations}. Stablecoins are a central node in these dynamics: they serve as settlement assets and collateral across protocols, and their de-pegs and runs can transmit stress through both on-chain and off-chain channels \cite{KosseEtAl2023Stablecoin,AnaduEtAl2023StablecoinsMMF,ESRB2025}. 
%yet, regulators also 
%and emphasized persistent gaps in quantitative tools that hinder monitoring and the construction of standardized risk indicators for the DeFi ecosystem. 
and underscored persistent gaps in quantitative tools that limit effective monitoring for the DeFi ecosystem \citep{BIS2025TVL}. This measurement gap is increasingly salient as links between crypto markets and the broader financial system deepen \cite{ESRB2025}.

\textbf{Network primitives.} A DeFi \emph{protocol} is a smart-contract application providing a financial service such as lending, automated market making, or liquid staking \citep{Schar2021DeFi,gogol2024sokdecentralizedfinancedefi}. Each protocol holds an inventory of \emph{tokens}, fungible on-chain assets whose USD value aggregates into Total Value Locked (TVL) \citep{BIS2025TVL,DefiLlama2026}. Because tokens issued by one protocol are routinely held by another, the cross-protocol state at each timestep forms a directed weighted graph \citep{wu2025dexposuredatasetbenchmarksinterprotocol}: nodes are protocols, edges are token-mediated USD exposures, and node features are tabular descriptors such as TVL, token composition, and sector. The \emph{credit exposure} of $p$ to $q$ is the value of $q$-issued tokens held by $p$, equivalently the weight of edge $p\!\to\!q$; \emph{systemic risk} corresponds to shock propagation along multi-hop paths in this graph as nodes, edges, and weights evolve weekly \citep{EisenbergNoe2001,BattistonEtAl2012DebtRank,aufiero2025mappingmicroscopicsystemicrisks}.

Despite the importance, empirical work has been constrained by limited data and tools. %, leaving a gap for dynamic, ecosystem-wide measurement that can track how exposure concentrations and transmission channels evolve over time \cite{wu2025dexposuredatasetbenchmarksinterprotocol}.
DeFi exposures are not reported in standardized balance sheets; they must be reconstructed from on-chain states and transactions. Even when exposures can be estimated, they are (1) \emph{high-dimensional}: there are millions of potential directed links, (2) \emph{nonstationary}, in terms of protocol upgrades, new assets, changing incentives, and (3) \emph{strongly heteroskedastic}: regime shifts are frequent between calm and crisis. Classical time-series tools such as vector autoregression (VAR) \citep{Sims1980} and low-rank factor methods \citep{StockWatson2002} provide strong baselines, but they suffer from restricted capacity to represent data features, and struggle to represent the coupled evolution of \emph{node-level states} (e.g., total locked value in a protocol), \emph{edge-level exposures} (e.g., who is economically linked to whom, and with what weight), and \emph{meso-structure} (e.g., sectoral organization and its rewiring under stress). Meanwhile, modern graph neural networks (GNNs) \citep{KipfWelling2017,VelickovicEtAl2018} and temporal graph models for dynamic interaction data \citep{rossi2020temporalgraphnetworksdeep} have shown impressive performance in wide areas%offer expressive inductive biases
; yet they are typically trained for a narrow task on a limited dataset, rather than serving as reusable infrastructure across multiple tasks.

To address this gap, we train {DeXposure-FM}, a domain-specific time-series graph foundation model for measuring and forecasting inter-protocol credit exposures in global decentralized financial networks. The model is trained on more than 43.7 million weekly inter-protocol exposure data entries from 2020--2025, stored in the {DeXposure} dataset, which constructs credit exposures across over 4,300 protocols on 602 blockchains and covers 24,300+ unique tokens \cite{wu2025dexposuredatasetbenchmarksinterprotocol}. DeXposure constructs inter-protocol exposures by linking protocol-level balance-sheet proxies (e.g., total value locked (TVL) \citep{BIS2025TVL}) to token- and sector-level structure, enabling a unified view of volumes, topology, and compositional risk. The dataset captures an evolving, weighted, directed multigraph whose nodes are protocols and whose edges represent economically meaningful exposure links inferred from token composition and valuation dynamics. DeXposure-FM follows the foundation-model paradigm: it adapts a reusable graph-tabular representation to large-scale DeFi data and exposes it through task heads that serve forecasting and downstream measurement tasks \citep{BommasaniEtAl2021FoundationModels}.

DeXposure-FM employs a GraphPFN graph-tabular encoder \citep{eremeev2025graphpfn} to extract representations from input data, followed by multiple heads
%builds on a pre-trained GraphPFN encoder that jointly ingests each exposure-graph snapshot and protocol-level tabular descriptors, and attaches lightweight prediction heads 
to forecast (1) edge existence and edge weights on the graph, and (2) node-level TVL changes at multiple horizons. In optimization, the encoder is initialized from open-source GraphPFN weights and fine-tuned end-to-end together with randomly initialized task heads. Time-based walk-forward splits \citep{BergmeirBenitez2012CVTS}, early stopping \citep{Prechelt1998EarlyStopping}, and gradient clipping \citep{PascanuEtAl2013RNNTraining} are employed to improve robustness under non-stationary environments.
%\paragraph{Macroprudential objects on dynamic graphs}
In training, %The modeling challenge is not merely to predict edge weights; it is 
DeXposure-FM is optimized by Adam \citep{KingmaBa2015Adam} for the joint tasks of (1) edge existence classification, (2) edge weight regression, and (3) node-level TVL-change prediction. These outputs map naturally to policy-relevant statistics: concentration, density, sector connectivity, and directional spillovers. In crises, the relevant question is not only ``what is the expected TVL tomorrow?'', but ``which parts of the network amplify shocks, and along which pathways does distress travel?'' as well.

We evaluate DeXposure-FM on two machine learning benchmarks, tailored for economic measurement and risk analytics: (1) multi-step forecasting of edge-level exposures and network-level statistics (e.g., concentration, density, and sector connectivity), and (2) predictive stress testing, where we forecast a future exposure network and compare simulated system losses on predicted versus realized graphs under fixed counterfactual shock definitions. Across tasks, we benchmark against strong competitors, including GraphPFN, ROLAND \citep{YouEtAl2022ROLAND} (a temporal graph neural network), and a persistence baseline that carries forward the last observed graph structure.

%On the financial economics front, 
DeXposure-FM provides a range of financial economics tools for macroprudential monitoring in decentralized financial infrastructure. We use the trained model to construct dynamic measures of protocol-level systemic importance, sector-to-sector spillover measures (spillover matrix and spillover concentration index), and early-warning indicators based on shifts in network concentration and dependence on fragile collateral. These model-based measurements support DeFi-specific stress testing under counterfactual shocks, while clarifying when forecast-then-measure outputs are robust and when they require cautious interpretation by regulators and central banks. The empirical results show that learned forecasts are most useful in structural-change and tail-error regimes where the carry-forward assumption is least reliable.
%By combining novel on-chain data with a multimodal foundation model, 
As a domain-specific time-series graph foundation model for measuring and forecasting inter-protocol credit exposure on DeFi networks, DeXposure-FM advances the measurement toolkit needed to understand systemic risk in DeFi and its interaction with the wider financial system.

DeXposure-FM will be continuously updated and extended along five directions: (1) enlarging the training data pool; (2) improving exposure measurement by moving beyond raw TVL; (3) addressing the continuous model drift through periodic retraining on new data, drift monitoring by sector/chain, and versioned releases tied to specific data vintages; (4) innovating the architecture; and (5) fostering an open-source community by maintaining public code and model weights, running competitions, etc.

\section{Related Works}

This section reviews the literature in relevant areas.

\subsection{Credit Risk Modeling in Financial Markets}
In traditional finance, credit risk modeling spans both individual-default models and systemic network approaches.  Classical structural models and reduced-form models quantify default probabilities of single entities, while network contagion models examine how defaults propagate through financial institutions.  For example, Dolfin et al. apply network epidemic models to credit contagion, highlighting how interconnected portfolios can amplify systemic credit risk \cite{math7080713}.  
In decentralized finance, credit risk manifests differently, as loans are typically over-collateralized and mediated by smart contracts. Bertomeu et al. propose simple aggregate risk measures for DeFi lending protocols, using only total deposits and borrowings, and report that systemic fragility surged around mid-2021 during crypto market turmoil \cite{BERTOMEU2024105321}. Conceptual studies have begun to map TradFi and DeFi risks together: Aufiero et al. review systemic risk mechanisms, finding that while basic risk types (leverage, liquidity shocks, correlated exposures) are common to both systems, DeFi's algorithmic execution and composability lead to unique propagation channels \cite{aufiero2025mappingmicroscopicsystemicrisks}.

\subsection{Machine Learning for Systemic Risk, Contagion, and Network Analysis}
Machine learning techniques are increasingly applied to systemic risk detection and contagion modeling \citep{BALMASEDA2023200240,gonon2025computingsystemicriskmeasures}. Graph neural networks (GNNs) in particular leverage network structure to improve risk prediction \citep{KipfWelling2017,VelickovicEtAl2018}. Balmaseda et al. show that a GNN-based model dramatically outperforms traditional ML for classifying systemic importance of banks in simulated networks \cite{BALMASEDA2023200240}. Gonon et al. integrate GNNs with explicit interbank liability networks, computing systemic risk measures as the minimum capital needed to secure the system \cite{gonon2025computingsystemicriskmeasures}. Their framework effectively learns contagion channels (e.g., Eisenberg-Noe clearing) by incorporating graph-structured data into the risk aggregation function. %These modern ML approaches build on network science foundations.  %For instance, \cite{math7080713} emphasize that network-theoretic methods are crucial for modeling correlated systemic credit risk.  
%Overall, machine-learning-driven network analysis enables end-to-end forecasting of systemic events, aiming to detect contagion pathways and provide early warnings for financial crises.

\subsection{Foundation Models in Finance and Economics}

Foundation models are increasingly penetrating into financial and economic domains by shifting the workflow from task-specific tools toward reusable representations learned from broad data. Instead of fitting a separate model for each target, recent approaches pretrain large neural networks, mostly using an architecture of transformer \citep{VaswaniEtAl2017Attention} and its variants, on massive data; and then transfer the pre-trained model across domains and tasks via prompting or light adaptation \citep{AnsariEtAl2024Chronos,RasulEtAl2023LagLlama,DasEtAl2024TimesFM}. In parallel, an emerging policy-facing literature uses large language models to extract quantitative signals from narrative text (e.g., reports, releases, and news) and incorporate them into nowcasting pipelines, suggesting that relatively small but information-dense text sources can improve real-time prediction when combined with standard indicators \citep{deBondtSun2025ChatGPTNowcast,KwonEtAl2024BISLLMprimer,CarrieroEtAl2024MacroLLM}. %Beyond ``pure'' time series, foundation-model architectures for structured data broaden the toolkit for financial applications: \fh{\cite{Ying2021Graphormer,HollmannEtAl2025TabPFN}}
These models are also adapted to graph inputs, which augment self-attention with graph structural encodings, demonstrating the feasibility of large pretrained models on networks that resemble webs of financial contracts or cross-asset relationships; a good example is Graphormer \citep{Ying2021Graphormer}. For tabular data, the Tabular Prior-data Fitted Network (TabPFN) \citep{HollmannEtAl2025TabPFN} is pretrained on millions of synthetic tabular tasks and reports strong out-of-the-box performance on diverse benchmarks. GraphPFN \citep{eremeev2025graphpfn} extends the prior-data fitted network paradigm to attributed graphs by jointly encoding tabular node features and graph structure in a single transformer; a related line of work shows that tabular foundation models can themselves be turned into graph foundation models \citep{eremeev2025turningtabularfoundationmodels}. We use the open-source GraphPFN weights as initialization and fine-tune on DeFi. %with up to 10,000 samples, while also supporting generative uses such as density estimation and data generation. %Taken together, these advances imply that future DeFi risk models can integrate heterogeneous modalities; for example, coupling graph encoders for protocol exposure networks with transformer encoders over protocol- and token-level covariates, toward unified ``world models'' of financial ecosystems that forecast dynamics while aiming for robustness under regime shifts and measurement noise.

%\subsection{Graph and Tabular Foundation Models in Finance/DeFi}
%Emerging foundation-model architectures for graph and tabular data hold promise for financial applications.  Transformer-based models have been adapted to graph inputs: Graphormer \cite{Ying2021Graphormer} incorporates graph structural encodings (e.g., spatial and centrality features) into the Transformer, achieving state-of-the-art performance on a variety of graph prediction tasks.  This demonstrates the feasibility of large pretrained models on graph data, which could include networks of financial contracts or asset relationships.  For tabular data, recent work introduces powerful foundation models as well.  \cite{Hollmann2025} present the Tabular Prior-Data Fitted Network (TabPFN), a transformer model pretrained on millions of synthetic tabular tasks; TabPFN ``outperforms all previous methods'' on diverse tabular benchmarks with up to 10k samples.  As a generative model, TabPFN also supports fine-tuning, density estimation, and data generation, highlighting its versatility.  The development of such graph and tabular foundation models implies that future DeFi risk models can integrate heterogeneous data: for example, combining graph neural networks on protocol networks with transformer encoders on protocol-level features.  These advances provide a rich toolkit for building unified world models of financial ecosystems that operate across structured graph data and tabular features.

\section{Formalizing Credit-Exposures on Decentralized Financial Networks}

%In decentralized finance, protocols lock and issue tokens that circulate across multiple platforms.  When a lending protocol such as MakerDAO accepts wrapped ether (WETH) as collateral or a liquidity pool holds stablecoins minted by Circle or Tether, a web of implicit credit links arises: the value of one protocol’s liabilities depends on the solvency and liquidity of another. 
This section formally defines terminologies and notations. %Traditional financial network models highlight that interconnectedness can diversify risk but also create channels for contagion; the amplification or dampening of shocks depends on network structure, leverage, and recovery rates \cite{Glasserman2016}.  Early systemic‑risk research used clearing vectors to characterise interbank obligations and showed that fixed‑point algorithms can compute consistent payment outcomes even in the presence of cyclical interdependence \cite{EisenbergNoe2001}.  More recent work demonstrates that as networks become more complete, the stability of the system initially improves but may deteriorate beyond a critical threshold because dense interconnections propagate shocks \cite{Acemoglu2013}.  These insights motivate a rigorous formalisation of credit‑exposure in DeFi networks.

\subsection{Credit Exposure in DeFi Networks}

Let $X_G(q)$ denote the set of all tokens issued or generated by protocol $q$, and let $X_p(t)$ denote the set of tokens held (or locked) by protocol $p$ at time $t$.  We call that protocol $p$ has {credit exposure} to protocol $q$ at time~$t$ if the tokens it holds include at least one that was issued by $q$:
\begin{equation}
X_p(t)\cap X_G(q)\neq \varnothing.
\label{eq:credit-exposure-definition}
\end{equation}
In other words, $p$ holds a liability of $q$; if the tokens issued by $q$ lose value or become illiquid, then $p$ faces a corresponding loss.  This token‑mediated dependence generalizes the notion of counterparty credit risk in traditional finance to decentralized platforms, where exposures arise not from explicit bilateral contracts but from the possession of tokenized claims.  The dynamics of these relationships can be measured by reconstructed total value locked (TVL) time series for each protocol and tracking how the composition of tokens held by each protocol evolves over time \cite{wu2025dexposuredatasetbenchmarksinterprotocol}.

%\subsection{Illustrative example}

\begin{figure}[htbp]
    \centering
    \begin{tikzpicture}[node distance=2.5cm, auto, scale=0.85, transform shape]
        \tikzstyle{entity} = [rectangle, draw, fill=Apricot, minimum width=2.2cm, minimum height=0.9cm, font=\small]
        \tikzstyle{balance} = [rectangle, draw, fill=SpringGreen, minimum width=3.2cm, minimum height=2.4cm, font=\footnotesize]
        \tikzstyle{exposure} = [->, thick, dashed, red!70!black]

        % User at top
        \node[entity] (user) at (0,3) {User};

        % Three protocols in a row
        \node[entity] (lido) at (-5,0) {Lido};
        \node[entity] (wrapper) at (0,0) {wstETH Wrapper};
        \node[entity] (pendle) at (5,0) {Pendle};

        % Balance sheets below each protocol
        \node[balance] (balanceLido) at (-5,-2.8) {
            \begin{tabular}{lr}
                \textbf{Lido} & \\
                \underline{Assets} & \\
                Staked ETH & \$100 \\
                \underline{Liabilities} & \\
                stETH & \$100
            \end{tabular}
        };
        \node[balance] (balanceWrapper) at (0,-2.8) {
            \begin{tabular}{lr}
                \textbf{wstETH Wrapper} & \\
                \underline{Assets} & \\
                stETH & \$100 \\
                \underline{Liabilities} & \\
                wstETH & \$100
            \end{tabular}
        };
        \node[balance] (balancePendle) at (5,-2.8) {
            \begin{tabular}{lr}
                \textbf{Pendle} & \\
                \underline{Assets} & \\
                wstETH & \$100 \\
                \underline{Liabilities} & \\
                PT-wstETH & \$95 \\
                YT-wstETH & \$5
            \end{tabular}
        };

        % Token flows (solid arrows)
        \draw[->, thick] (user) -- node[left, font=\footnotesize] {ETH} (lido);
        \draw[->, thick] (lido) -- node[above, font=\footnotesize] {stETH} (wrapper);
        \draw[->, thick] (wrapper) -- node[above, font=\footnotesize] {wstETH} (pendle);
        \draw[->, thick] (pendle) to[bend left=20] node[right, font=\footnotesize] {PT + YT} (user);

        % Credit exposure arrows (dashed red)
        \draw[exposure] (pendle.south west) to[bend right=15] node[below, font=\footnotesize, text=red!70!black] {exposure} (wrapper.south east);
        \draw[exposure] (wrapper.south west) to[bend right=15] node[below, font=\footnotesize, text=red!70!black] {exposure} (lido.south east);

    \end{tikzpicture}
    \caption{Multi-layer credit exposure in DeFi. A user stakes ETH with Lido, receiving stETH, which is wrapped into wstETH and deposited into Pendle. Each protocol holds the liability of the preceding one, creating a chain of credit exposures from Pendle to Lido.}
    \label{fig:exposure_example}
\end{figure}

Figure~\ref{fig:exposure_example} illustrates how a single-user transaction creates a multi-layer chain of credit exposures in DeFi.  A user deposits ETH into Lido, a liquid staking protocol, and receives stETH, a rebasing token that accrues staking rewards.  The stETH is then wrapped into wstETH, a non-rebasing representation that is more compatible with DeFi protocols.  Finally, wstETH is deposited into Pendle, a yield-tokenization protocol that splits the asset into a principal token PT-wstETH and a yield token YT-wstETH, allowing users to trade fixed and variable yield separately.

In balance-sheet terms, Lido holds staked ETH with validators and issues stETH as liabilities, the wstETH wrapper holds stETH and issues wstETH, and Pendle holds wstETH and issues the PT and YT pair.  Because each protocol's assets consist of tokens issued by the preceding protocol, a chain of credit exposures emerges: Pendle has exposure to the wstETH wrapper, which in turn has exposure to Lido.  If Lido experienced a slashing event or stETH de-pegged from ETH, the losses would cascade through wstETH and into Pendle, affecting PT and YT holders.  This example demonstrates that DeFi exposures are inherently multi-layered, as liquid staking, wrapping, and yield tokenization compound into complex dependency structures that the DeXposure dataset aims to capture.

\subsection{Mapping Tokens to Issuing Protocols}

To operationalize eq. \eqref{eq:credit-exposure-definition}, we must determine which protocol issues each token and thus identify the direction of exposure.  Let $\mathcal{P}$ denote the set of all protocols and $\mathcal{C}$ the set of blockchains.  For each token $\sigma$ appearing in protocol $p$, a {mapping function} $M(\sigma)$ returns the protocol $q$ that issues or manages $\sigma$.  We can construct $M$ using a fallback procedure:
\begin{enumerate}
  \item \emph{Metadata lookup:} Tokens are directly matched to issuing protocols using metadata from DefiLlama \citep{DefiLlama2026} when available.
  \item \emph{Manual mapping:} For tokens with high TVL that lack metadata links, researchers manually assign the issuing protocol based on documentation and expert knowledge.
  \item \emph{Text‑vector similarity:} For remaining tokens, the text descriptions of tokens and protocols are vectorized via the term frequency–inverse document frequency (TF-IDF) representation \cite{Qaiser2018}.  Cosine similarity scores between token and protocol vectors determine the best match, and tokens are mapped to the protocol with the highest similarity score exceeding a threshold $\theta$.
  \item \emph{Primary market tokens:} If none of the above methods yields a mapping, the token is treated as its own protocol (e.g., WETH).
\end{enumerate}
This mapping ensures that token flows between protocols can be attributed to changes in credit exposure with respect to the correct issuing protocol, even when tokens are bridged across chains or wrapped by multiple platforms. This procedure was implemented to curate the DeXposure dataset.

\subsection{Network Representation and Value Flows}

Credit exposures over a discrete time interval $\tau = t_2 - t_1$ are represented by a weighted directed graph $G_\tau = (P_\tau, E_\tau)$, where $P_\tau$ is the set of active protocols and $E_\tau$ is the set of directed edges.  Each vertex $p\in P_\tau$ is assigned a {node weight} equal to the value of assets it holds at the end of the interval:
\begin{equation}
w_\tau(p) = \sum_{\sigma\in X_p(t_1)\cap X_p(t_2)} v_{t_2,\sigma},
\label{eq:node-weight}
\end{equation}
where $v_{t_2,\sigma}$ is the USD value of token $\sigma$ at time $t_2$.  Nodes with negligible weight (below a threshold $\theta$) can be pruned for clarity.  To define the {edge weight} from protocol $p$ to protocol $q$, we first compute the value flow for each token $\sigma$:
\begin{equation}
F^{\sigma}_{p q,\tau} =
\begin{cases}
\max\bigl(0,\, -\Delta S^{\sigma}_{p,\tau}\bigr), & \text{if } \Delta S^{\sigma}_{p,\tau} < 0,\\
\max\bigl(0,\,\Delta S^{\sigma}_{q,\tau}\bigr), & \text{if } \Delta S^{\sigma}_{q,\tau} \ge 0,
\end{cases}
\label{eq:value-flow}
\end{equation}
where $\Delta S^{\sigma}_{p,\tau}=v_{p,t_2}^{\sigma}-v_{p,t_1}^{\sigma}$ is the change in the USD value of token $\sigma$ held by protocol $p$ over the interval.  Eq.~\eqref{eq:value-flow} reflects the intuition that if $p$ decreases its holdings of $\sigma$ and $q$ increases theirs, then $\sigma$ has flowed from $p$ to $q$; negative flows are reversed so that all flows are non‑negative.  The {edge weight} $w_\tau(e_{p q})$ is the sum of flows for all tokens that map to the issuing protocol $q$:
\begin{equation}
w_\tau(e_{p q}) = \sum_{\sigma \in X_{p q, \tau}} F^{\sigma}_{p q,\tau},
\label{eq:edge-weight}
\end{equation}
where $X_{p q,\tau}$ denotes the set of tokens with issuing protocol $q$ that move from $p$ to $q$ over $\tau$.  A positive edge weight indicates increasing exposure from $p$ to $q$.  For single‑token protocols, the flow of that token equals the edge weight.

%Algorithm~1 in \cite{wu2025dexposuredatasetbenchmarksinterprotocol} describes a procedure for constructing $G_\tau$ from raw token‑protocol‑chain mappings.  
In summary, one aggregates token holdings at the start and end of the interval, computes node weights, discards nodes below a threshold, and then computes edge weights by summing positive flows; negative flows are reversed to ensure all weights are non‑negative.  The resulting time‑indexed sequence of graphs captures the evolving structure of credit dependencies in the DeFi ecosystem. The curation of the DeXposure dataset follows this setting.

\section{DeXposure-FM: Architecture and Optimization}

This section introduces the DeXposure-FM, including its architecture and the training process. %The trained model is available at \url{https://huggingface.co/EVIEHub/DeXposure-FM}. The code is released at \url{https://github.com/EVIEHub/DeXposure-FM}.

\subsection{Model Architecture}\label{subsec:model-arch}

DeXposure-FM is a time-series graph foundation model designed to forecast the evolution of inter-protocol credit exposures. The architecture follows an \emph{encoder-multi-head} design: a pretrained graph-tabular encoder produces protocol embeddings for each weekly snapshot, and task heads map these embeddings to (1) edge existence and edge-weight residuals and (2) node-level TVL changes. Network-level statistics are then obtained by applying deterministic functionals to the predicted graph.

\subsubsection{Input and Features}
Let $t\in\mathcal{T}$ be discrete observation times and $\tau=(t_1,t_2)$ be a weekly interval. For each $\tau$, we construct a weighted directed exposure graph
\begin{equation}
G_\tau=(\mathcal{P}_\tau,\mathcal{E}_\tau),
\end{equation}
where nodes $p\in\mathcal{P}_\tau$ are protocols and edges $(p,q)\in\mathcal{E}_\tau$ represent token-mediated exposure from $p$ to issuing protocol $q$. Node weights encode protocol size (TVL) and edge weights encode the weekly change in exposure induced by token flows and valuation updates. Each node $p$ additionally has a tabular descriptor vector $ x^{\text{tab}}_{p,\tau}$ including log-scaled TVL and token-composition summaries (e.g., number of token types held, concentration measures, entropy) and a sector/category one-hot encoding.

\subsubsection{Pretrained Graph-Tabular Encoder}

We adopt {GraphPFN} as the encoder $E(\cdot)$ in our model, which is a transformer-based graph-tabular foundation model that fuses graph structure with tabular node features via multi-head self-attention augmented with structural information \citep{eremeev2025graphpfn}. Originally, GraphPFN is trained as a prior-data fitted network for node-level classification and regression on attributed graphs. In DeXposure-FM, we discard GraphPFN's original node-prediction head, initialize the encoder from the open-source weights, and fine-tune it end-to-end with the new multi-task heads. Given $(G_\tau,\{ x^{\text{tab}}_{p,\tau}\}_{p\in\mathcal{P}_\tau})$, the encoder produces
$d$-dimensional node representations
\begin{equation}
\{h_{p,\tau}\}_{p\in\mathcal{P}_\tau}
= E_{\text{GraphPFN}}\!\left(G_\tau,\{ x^{\text{tab}}_{p,\tau}\}_{p\in\mathcal{P}_\tau}\right),
\qquad h_{p,\tau}\in\mathbb{R}^d .
\end{equation}

\subsubsection{Edge-Level Prediction Head}

For each forecast horizon $h\in\{1, 4, 8, 12\}$ weeks, DeXposure-FM predicts both edge existence and edge weight (i.e., exposure change) for ordered protocol pairs $(p,q)$. We form a pairwise feature vector using a standard symmetric composition of embeddings:
\begin{equation}
 f_{pq,\tau}=\Big[ h_{p,\tau};\,  h_{q,\tau};\,  h_{p,\tau}\odot   h_{q,\tau};\,\lvert  h_{p,\tau}-  h_{q,\tau}\rvert\Big],
\end{equation}
and pass it through a multi-layer perceptron (MLP) head to obtain
\begin{equation}
(\widehat{y}^{\mathrm{exist}}_{pq}, \widehat{r}_{pq,\tau,h}) = \mathrm{MLP}_{\mathrm{link}}({f}_{pq,\tau}),
\end{equation}
where $\widehat{y}^{\mathrm{exist}}_{pq}\in(0,1)$ is the predicted probability of edge existence and $\widehat{r}_{pq,\tau,h}\in\mathbb{R}$ is a residual in log space. In our implementation we adopt residual learning for edge weights: letting $\tilde{w}_{\tau}(e_{pq})=\log(1+w_{\tau}(e_{pq}))$, we reconstruct the predicted log-weight as
\begin{equation}
\widehat{\tilde{w}}_{\tau+h}(e_{pq})=\tilde{w}_{\tau}(e_{pq})+\widehat{r}_{pq,\tau,h},
\end{equation}
with $\tilde{w}_{\tau}(e_{pq})=0$ when the edge is absent at time $\tau$.
\subsubsection{Node-Level Prediction Head and Network Statistics}

To capture protocol-level dynamics, we predict the log-change in protocol TVL using a node head:
\begin{align*}
    \widehat{\Delta}_{p,\tau+h} = &\mathrm{MLP}_{\mathrm{node}}\!\Big([h_{p,\tau};\,h^{\mathrm{in}}_{p,\tau};\,h^{\mathrm{out}}_{p,\tau}]\Big),\\
\widehat{\Delta}_{p,\tau+h} =& \log(1+w_{\tau+h}(p)) - \log(1+w_{\tau}(p)),
\end{align*}
for protocols $p\in\mathcal{P}_{\tau}\cap\mathcal{P}_{\tau+h}$.
Here $h^{\mathrm{in}}_{p,\tau}$ and $h^{\mathrm{out}}_{p,\tau}$ denote simple weighted aggregations of incoming/outgoing neighbor embeddings at time $\tau$ (using the current-week edge weights), which inject local flow context into the TVL-change prediction.

Network-level statistics (e.g., density, concentration measures, and sector connectivity) are computed as deterministic functionals of the predicted graph $\widehat{G}_{\tau+h}$. This separation lets the same encoder and heads serve both forecasting targets and the macroprudential measurement layer in Section~\ref{sec:econ}.

\subsection{Training Pipeline}\label{subsec:training-pipeline}

This section describes how we construct supervised training pairs from weekly DeXposure snapshots, define the multi-task learning objective, and optimize DeXposure-FM.

\subsubsection{Data Preparation}\label{subsubsec:data-prep}
DeXposure provides weekly snapshots of the inter-protocol exposure network. For each discrete time index $\tau$, we construct a weighted directed graph
$G_{\tau}=(\mathcal{P}_{\tau},\mathcal{E}_{\tau})$ with node weights $w_{\tau}(p)$ (protocol TVL) and edge weights $w_{\tau}(e_{pq})$ (exposure from $p$ to $q$).
Training examples are formed by an \emph{anchor} snapshot and a forecast horizon:
\begin{equation}
G_{\tau} \;\longrightarrow\; G_{\tau+h},
\qquad
h\in\{1,4,8,12\},
\label{eq:train-pairs}
\end{equation}
so that the model predicts the future graph state at horizon $h$ conditional on the current snapshot.

For each protocol $p\in\mathcal{P}_{\tau}$ we provide tabular node features $x^{\text{tab}}_{p,\tau}$ (including log-scaled TVL, token counts, concentration measures, and sector/category indicators) together with the graph structure $G_{\tau}$.

Table~\ref{tab:data-stats} summarizes the dataset characteristics. The network exhibits high edge persistence (mean overlap ratio 98.5\%), reflecting the stable nature of DeFi protocol interconnections.

\begin{table}[t]
\centering
\small
\begin{tabular}{lr}
\hline
\textbf{Statistic} & \textbf{Value} \\
\hline
Total snapshots & 283 weeks \\
Time range & 2020-03-23 $\sim$ 2025-08-18 \\
Mean nodes per week & 5,676 \\
Mean edges per week & 30,424 \\
Mean edge overlap ratio & 98.5\% \\
Protocol categories & $\sim$15 classes \\
\hline
\end{tabular}
\caption{DeXposure dataset statistics.}
\label{tab:data-stats}
\end{table}

\textbf{Time-based expanding window splits.}
We adopt an expanding-window walk-forward evaluation protocol to prevent look-ahead bias: all validation and test targets occur strictly after the corresponding training window, and the training window expands over time while validation and test windows remain fixed.

\textbf{Negative sampling.}
Edge existence is highly imbalanced.  For each horizon $h$, we construct a labeled edge set $\mathcal{S}_{\tau+h}$ consisting of all positive edges in $\mathcal{E}_{\tau+h}$ plus uniformly sampled negative edges from the complement, using a negative-to-positive ratio of $5{:}1$.
This sampled set is used for the edge existence loss, while the edge weight loss is computed only on positive edges.

\subsubsection{Loss Functions}\label{subsubsec:loss}
DeXposure-FM is trained with a multi-task objective combining (1) edge existence classification, (2) edge weight regression, and (3) node-level TVL-change prediction.

\textbf{Edge existence (link prediction).}
For each candidate pair $(p,q)\in\mathcal{S}_{\tau+h}$ we predict $\widehat{y}^{\text{exist}}_{pq}\in(0,1)$ and minimize a weighted binary cross-entropy loss:
\begin{equation}
\mathcal{L}_{\text{exist}}
=
\sum_{(p,q)\in\mathcal{S}_{\tau+h}}
\mathrm{BCE}\!\left(\widehat{y}^{\text{exist}}_{pq},\,y^{\text{exist}}_{pq};\,w_{\text{pos}}\right),
\label{eq:loss-exist}
\end{equation}
where $w_{\text{pos}}$ upweights positives to match the negative sampling ratio. Here, binary cross-entropy is defined as below. 

\begin{definition}[binary cross-entropy (BCE)]
For a binary label $y\in\{0,1\}$ and a predicted probability $\hat{y}\in(0,1)$, the binary cross-entropy loss is
\begin{equation}
\mathrm{BCE}(\hat{y},y)
= -\Big(y\log(\hat{y}) + (1-y)\log\big(1-\hat{y}\big)\Big).
\label{eq:bce}
\end{equation}
Equivalently, if $\hat{y}=\sigma(z)$ is produced from a logit $z\in\mathbb{R}$ via the sigmoid $\sigma(z)=1/(1+e^{-z})$, then
\begin{equation}
\mathrm{BCE}(\sigma(z),y)
= \log\!\big(1+e^{z}\big) - yz,
\label{eq:bce-logit}
\end{equation}
which is numerically stable and is often referred to as the logistic loss.
\end{definition}

\textbf{Edge weights (residual learning).}
For positive edges $(p,q)\in\mathcal{E}_{\tau+h}$ we predict a residual in log space relative to the previous observed week. Let $\tilde{w}_{\tau}(e_{pq})=\log(1+w_{\tau}(e_{pq}))$ and $\tilde{w}_{\tau+h}(e_{pq})=\log(1+w_{\tau+h}(e_{pq}))$. The link head outputs $\widehat{r}_{pq,\tau,h}$ and we apply a robust Smooth~L1 loss:

\begin{equation}
\mathcal{L}_{\text{weight}}
=
\sum_{(p,q)\in\mathcal{E}_{\tau+h}}
\mathrm{SmoothL1}\!\Big(
\widehat{r}_{pq,\tau,h}
-
\big(\tilde{w}_{\tau+h}(e_{pq})-\tilde{w}_{\tau}(e_{pq})\big)
\Big).
\label{eq:loss-weight}
\end{equation}

Equivalently, we reconstruct $\widehat{\tilde{w}}_{\tau+h}(e_{pq})=\tilde{w}_{\tau}(e_{pq})+\widehat{r}_{pq,\tau,h}$ and minimize $\mathrm{SmoothL1}(\widehat{\tilde{w}}_{\tau+h}(e_{pq})-\tilde{w}_{\tau+h}(e_{pq}))$. We set $\tilde{w}_{\tau}(e_{pq})=0$ when the edge is absent at time $\tau$. The $\log(1+\cdot)$ transform stabilizes training under the heavy-tailed exposure distribution. Here, smooth L1 (Huber) loss is defined as follows.

\begin{definition}[smooth L1 (Huber) loss]
For a scalar prediction $\hat{u}\in\mathbb{R}$ and target $u\in\mathbb{R}$, let the residual be $r=\hat{u}-u$.
The Smooth~L1 (Huber) loss with threshold $\delta>0$ is defined as
\begin{equation}
\mathrm{SmoothL1}_\delta(r)=
\begin{cases}
\frac{1}{2}\frac{r^2}{\delta}, & \text{if } |r|<\delta,\\[6pt]
|r|-\frac{1}{2}\delta, & \text{otherwise}.
\end{cases}
\label{eq:smoothl1}
\end{equation}
Equivalently, written directly in terms of $(\hat{u},u)$:
\begin{equation}
\mathrm{SmoothL1}_\delta(\hat{u},u)=
\begin{cases}
\frac{1}{2}\frac{(\hat{u}-u)^2}{\delta}, & \text{if } |\hat{u}-u|<\delta,\\[6pt]
|\hat{u}-u|-\frac{1}{2}\delta, & \text{otherwise}.
\end{cases}
\end{equation}
\end{definition}

Smooth~L1 behaves like an $\ell_2$ loss near zero (encouraging small residuals) and like an $\ell_1$ loss for large residuals (reducing sensitivity to heavy tails and outliers). A common choice is $\delta=1$, when $\mathrm{SmoothL1}_\delta$ is briefly written as $\mathrm{SmoothL1}$.

\textbf{Node-level TVL dynamics.}
We also predict the log-change in protocol TVL,
$\Delta_{p,\tau+h}=\log(1+w_{\tau+h}(p))-\log(1+w_{\tau}(p))$,
using Smooth~L1:
\begin{equation}
\mathcal{L}_{\text{node}}
=
\sum_{p\in\mathcal{P}_{\tau}\cap \mathcal{P}_{\tau+h}}
\mathrm{SmoothL1}\!\left(\widehat{\Delta}_{p,\tau+h}-\Delta_{p,\tau+h}\right),
\label{eq:loss-node}
\end{equation}
where $\Delta_{p,\tau+h} = \log(1+w_{\tau+h}(p)) - \log(1+w_{\tau}(p))$ is the ground-truth log-change in TVL.

\textbf{Multi-task objective.}
The total loss is a weighted sum:
\begin{equation}
\mathcal{L}
=
\lambda_{\text{exist}}\mathcal{L}_{\text{exist}}
+
\lambda_{\text{weight}}\mathcal{L}_{\text{weight}}
+
\lambda_{\text{node}}\mathcal{L}_{\text{node}},
\label{eq:loss-total}
\end{equation}
with $\lambda_{\text{exist}}=2.0$, $\lambda_{\text{weight}}=0.5$, and $\lambda_{\text{node}}=20.0$. These weights are calibrated to balance the gradient contributions across tasks despite their different loss magnitudes: edge existence loss $\mathcal{L}_{\mathrm{exist}} \approx 0.22$, edge weight loss $\mathcal{L}_{\mathrm{weight}} \approx 2.4$, and node loss $\mathcal{L}_{\mathrm{node}} \approx 0.05$ on average. They also reflect the primary importance of link prediction while using weight and node supervision to encourage economically meaningful representations. 

\subsubsection{Optimization}\label{subsubsec:optim}

\textbf{Initialization.} We discard GraphPFN's original node-prediction head, adopt the open-source GraphPFN weights as initialization of the encoder, and randomly initialize the edge and node heads used for DeXposure forecasting.

\textbf{Optimizer.}
We use Adam \citep{KingmaBa2015Adam} with hyperparameters $\beta_1=0.9$, $\beta_2=0.999$, learning rate $5 \times 10^{-4}$ for the task heads, and learning rate $5 \times 10^{-5}$ for the GraphPFN backbone during fine-tuning.

\textbf{Training details.}
We train for up to 20 epochs with early stopping (patience = 3) based on validation AUPRC. Gradient clipping ($\|\nabla\|_2 \le 1.0$) is applied for stability.

\textbf{Optional sharpness-aware training.}
Sharpness-Aware Minimization (SAM) is a promising extension to improve robustness under non-stationarity by favoring flatter optima. However, our current released training pipeline and the experiments reported in this paper use Adam; we leave SAM-based training as future work.

\begin{table}[t]
\centering
\small
\setlength{\tabcolsep}{6pt}
\renewcommand{\arraystretch}{1.05}
\begin{tabular}{ll}
\hline
\textbf{Hyperparameter} & \textbf{Value} \\
\hline
Data granularity & Weekly snapshots \\
Forecast horizons $h$ & $\{1, 4, 8, 12\}$ weeks \\
Negative sampling ratio & 5:1 (neg:pos) \\
Optimizer & Adam ($\beta_1=0.9$, $\beta_2=0.999$) \\
Learning rate (heads) & $5 \times 10^{-4}$ \\
Learning rate (backbone) & $5 \times 10^{-5}$ (fine-tune only) \\
Training epochs & 20 (early stopping patience=3) \\
Gradient clipping & $\|\nabla\|_2 \le 1.0$ \\
\hline
\multicolumn{2}{l}{\textit{Loss weights:}} \\
$\lambda_{\mathrm{exist}}$ (edge existence) & $2.0$ \\
$\lambda_{\mathrm{weight}}$ (edge weight) & $0.5$ \\
$\lambda_{\mathrm{node}}$ (node prediction) & $20.0$ \\
\hline
\end{tabular}
\caption{Training hyperparameters for DeXposure-FM experiments.}
\label{tab:training-hparams}
\end{table}

\noindent Empirical evaluations on machine-learning benchmarks are reported in Section~\ref{sec:evaluation}, and financial-economics tools and case-study verifications are presented in Section~\ref{sec:econ}.

\begin{comment}
    
\newpage

\section{Empirical Evaluations on Machine Learning Benchmarks}

\subsection{Task 1:  Multi-step Forecasting}

\textbf{Implementation Deatails}

\textbf{Empirical Results}

\subsection{Task 2: xx}

\textbf{Implementation Deatails}

\textbf{Empirical Results}

\section{Financial Economics Tools and Experimental Verifications}

\subsection{Macroprudential Applications}

\textbf{Systemic importance ranking.}

\textbf{Sector spillover monitoring. }

\textbf{Stress testing and contagion assessment.}

\textbf{Early warning signals.}

\subsection{Validity and Scope}
\end{comment}

\section{Empirical Evaluations on Machine Learning Benchmarks}
\label{sec:evaluation}

We evaluate DeXposure-FM on two machine learning benchmarks: (1) {multi-step forecasting} of edge existence, edge weights, and node TVL changes; and (2) {predictive contagion stress testing}, where we forecast future simulator-implied stress-test losses under fixed counterfactual shocks. The following Section~\ref{sec:econ} provides economic interpretation and applications of these predictive outputs.

\subsection{Baseline and Competitors}

We compare our model with three baselines:
(1) \textbf{GraphPFN}: a pre-trained GraphPFN encoder with frozen weights; only lightweight task-specific prediction heads are trained on DeXposure (a small MLP probe). In contrast, DeXposure-FM uses GraphPFN weights as initialization and fine-tunes the encoder and task heads end-to-end.
(2) \textbf{ROLAND}: a representative temporal graph neural network \citep{YouEtAl2022ROLAND}, trained from scratch.
(3) \textbf{Persistence}: a naive baseline that carries forward the last observed graph and edge weights ($\hat{A}_{t+h}=A_t$, $\hat{W}_{t+h}=W_t$).

For the trained baselines (GraphPFN and ROLAND), we use the same optimizer (Adam), learning rate ($5 \times 10^{-4}$), and training schedule as DeXposure-FM.

\subsection{Task I: Multi-step Forecasting}
\label{subsec:forecast}

We first evaluate DeXposure-FM on task of multi-step forecasting.

\subsubsection{Implementation Details}
\label{subsubsec:task1-impl}

We evaluate multi-step forecasts at horizons $h \in \{1,4,8,12\}$ weeks on three targets: (1) \emph{edge existence}, a binary classification task of whether an edge $(p,q)$ is present at $t+h$, (2) \emph{edge weight}, a regression task on log-scaled exposure weight for edges that exist, and (3) \emph{node TVL change}, a regression task. %on $\Delta_{p,t+h}=\log(1+\text{TVL}_{p,t+h})-\log(1+\text{TVL}_{p,t})$. 
We report AUROC \citep{Fawcett2006ROC} and AUPRC \citep{DavisGoadrich2006PR} for edge existence and MAE and RMSE for edge weights and $\Delta$TVL (all on the log scale).

\subsubsection{Empirical Results}
\label{subsubsec:task1-results}
%% [Updated to h={1,4,8,12} with complete DeXposure-FM results from GPU server experiments]

\begin{table}[t]
\centering
\small
\setlength{\tabcolsep}{4pt}
\begin{tabular}{llcccccc}
\hline
& & \multicolumn{2}{c}{\textbf{Edge Exist}} & \multicolumn{2}{c}{\textbf{Edge Weight}} & \multicolumn{2}{c}{\textbf{Node $\Delta$TVL}} \\
\textbf{Model} & $h$ & AUROC & AUPRC & MAE & RMSE & MAE & RMSE \\
\hline
\multirow{4}{*}{DeXposure-FM}
& 1 & \textbf{0.995} & \textbf{0.972} & \textbf{2.465} & \textbf{3.388} & \textbf{0.056} & \textbf{0.400} \\
& 4 & \textbf{0.995} & \textbf{0.973} & 2.489 & \textbf{3.424} & 0.140 & \textbf{0.680} \\
& 8 & \textbf{0.994} & \textbf{0.967} & \textbf{2.554} & \textbf{3.509} & \textbf{0.229} & \textbf{0.890} \\
& 12 & \textbf{0.993} & \textbf{0.967} & 2.648 & \textbf{3.606} & 0.286 & \textbf{1.046} \\
\hline
\multirow{4}{*}{GraphPFN (frozen)}
& 1 & 0.988 & 0.938 & 3.260 & 4.383 & 0.059 & 0.401 \\
& 4 & 0.988 & 0.940 & 3.189 & 4.049 & 0.142 & 0.682 \\
& 8 & 0.987 & 0.938 & 3.169 & 4.064 & 0.215 & 0.896 \\
& 12 & 0.986 & 0.936 & 3.136 & 4.130 & 0.324 & 1.056 \\
\hline
\multirow{4}{*}{ROLAND}
& 1 & 0.961 & 0.865 & 3.240 & 4.264 & 0.060 & 0.403 \\
& 4 & 0.962 & 0.868 & 3.242 & 4.162 & 0.141 & 0.684 \\
& 8 & 0.961 & 0.867 & 3.213 & 4.180 & 0.221 & 0.895 \\
& 12 & 0.961 & 0.866 & 3.195 & 4.177 & 0.279 & 1.058 \\
\hline
\multirow{4}{*}{Persistence}
& 1 & 0.763 & 0.604 & 2.487 & 4.296 & 0.057 & 0.403 \\
& 4 & 0.782 & 0.635 & \textbf{2.372} & 4.138 & \textbf{0.138} & 0.685 \\
& 8 & 0.749 & 0.580 & 2.618 & 4.400 & \textbf{0.213} & 0.899 \\
& 12 & 0.762 & 0.603 & \textbf{2.541} & 4.304 & \textbf{0.272} & 1.065 \\
\hline
\end{tabular}
\caption{Multi-step forecasting on the 2025 hold-out at horizons $h \in \{1,4,8,12\}$ weeks. DeXposure-FM = fine-tuned graph-tabular foundation model; GraphPFN (frozen) = pretrained encoder with a trainable MLP probe; Persistence carries forward $A_t,W_t$. All regression metrics are computed on log-scale exposures. Bold marks the best value in each metric and horizon.}
\label{tab:main-results}
\end{table}

Table~\ref{tab:main-results} reports the forecasting results on the strict 2025 hold-out test set. Despite the network's strong week-to-week persistence (98.5\% mean edge overlap), DeXposure-FM substantially improves \emph{edge existence} prediction: AUROC 0.993--0.995 and AUPRC 0.967--0.973 across horizons, versus AUPRC 0.58--0.64 for persistence and 0.936--0.940 for the frozen GraphPFN probe. On \emph{edge weights}, DeXposure-FM has lower errors than the learned neural baselines (GraphPFN-frozen and ROLAND); persistence remains competitive because last week's exposure magnitude is itself a strong predictor. On \emph{node $\Delta$TVL}, MAE is close to persistence at short horizons and slightly higher at longer horizons, while RMSE is lower than persistence across all four horizons. The pattern is consistent with the domain structure: graph-foundation representations are most useful for predicting edge formation and decay, while highly persistent magnitude targets leave less room over a carry-forward baseline.

\subsection{Task II: Predictive Contagion Stress Testing}
\label{subsec:stability}

We then evaluate our model on whether forward-looking network forecasts translate into accurate forecasts of \emph{stress-test losses} generated by a fixed contagion simulator. Specifically, we compare system losses computed on the model-predicted future network $\hat{G}_{t+h}$ with losses on the realized future network $G_{t+h}$, under fixed counterfactual shock definitions.

\subsubsection{Implementation Details}
\label{subsubsec:task2-impl}

We train DeXposure-FM on the pre-2025 training period (Section~\ref{subsec:training-pipeline}) and evaluate on the strict 2025 hold-out set. Each test week provides an origin $t$, yielding multiple out-of-sample pairs $(t,t+h)$ for horizons $h\in\{1,4,8,12\}$. To handle protocol entry/exit, all comparisons are computed on the common node set $V_t \cap V_{t+h}$.

Stress-test losses are computed using a DebtRank-style contagion simulator \citep{EisenbergNoe2001,BattistonEtAl2012DebtRank} under three scenarios: (1) \emph{top protocol shock}: 50\% TVL loss to the largest protocol, (2) \emph{top-5 protocols shock}: 30\% TVL loss to the top-5 protocols, and (3) \emph{bridge sector shock}: 100\% TVL loss to bridge protocols. For each $(t,t+h)$, we run the same simulator on three graphs: the observed network at $t$ (persistence baseline), the predicted network $\hat{G}_{t+h}$, and the realized network $G_{t+h}$.

Because the DeFi exposure network is highly persistent (Table~\ref{tab:data-stats}), the persistence baseline is difficult to beat on average. We therefore report both (i) overall performance and (ii) a \emph{post-hoc stratified evaluation} on the \textbf{worst 20\%} of test cases under persistence, defined by the largest baseline absolute errors within each horizon (pooled across shock scenarios). This stratification is computed on the held-out test set and is not used for model selection. We summarize improvement with $\Delta \mathrm{MAE}=\mathrm{MAE}(\text{baseline})-\mathrm{MAE}(\text{model})$, where positive values indicate lower error than persistence.

\subsubsection{Empirical Results}
\label{subsubsec:task2-results}
\begin{figure}[t]
\centering
\includegraphics[width=0.95\textwidth]{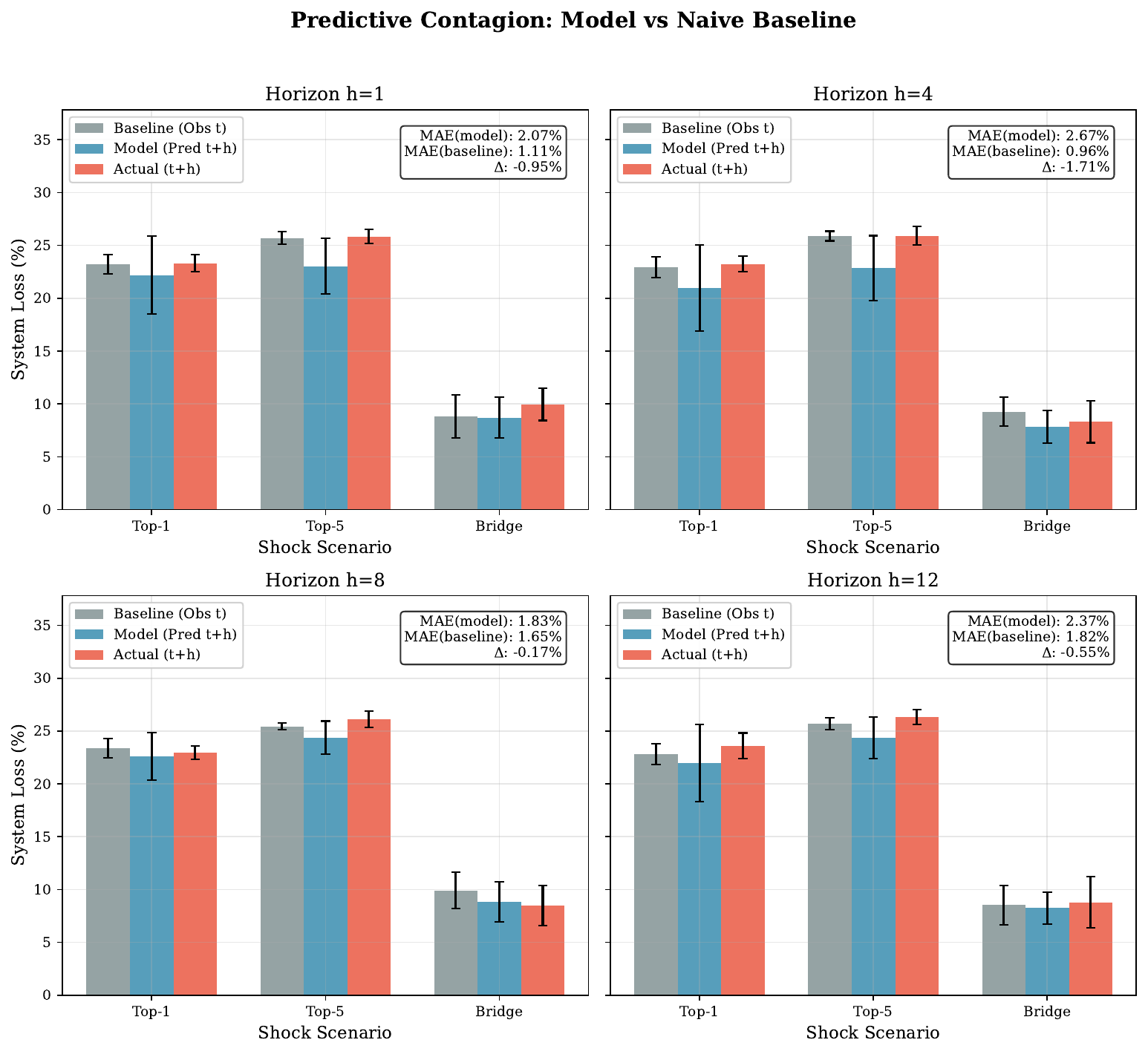}
\caption{Three-way predictive contagion stress testing: system-loss percentage under identical shock scenarios on the persistence graph $G_t$, the model forecast $\hat{G}_{t+h}$, and the realized graph $G_{t+h}$. Because DeFi exposure networks have 98.5\% week-to-week edge overlap, the persistence graph is itself a near-perfect input to the simulator on the largest exposure links that drive contagion; the model bars accordingly sit below the realized bars on the full sample.}
\label{fig:contagion-comparison}
\end{figure}

\begin{figure}[t]
\centering
\includegraphics[width=0.95\textwidth]{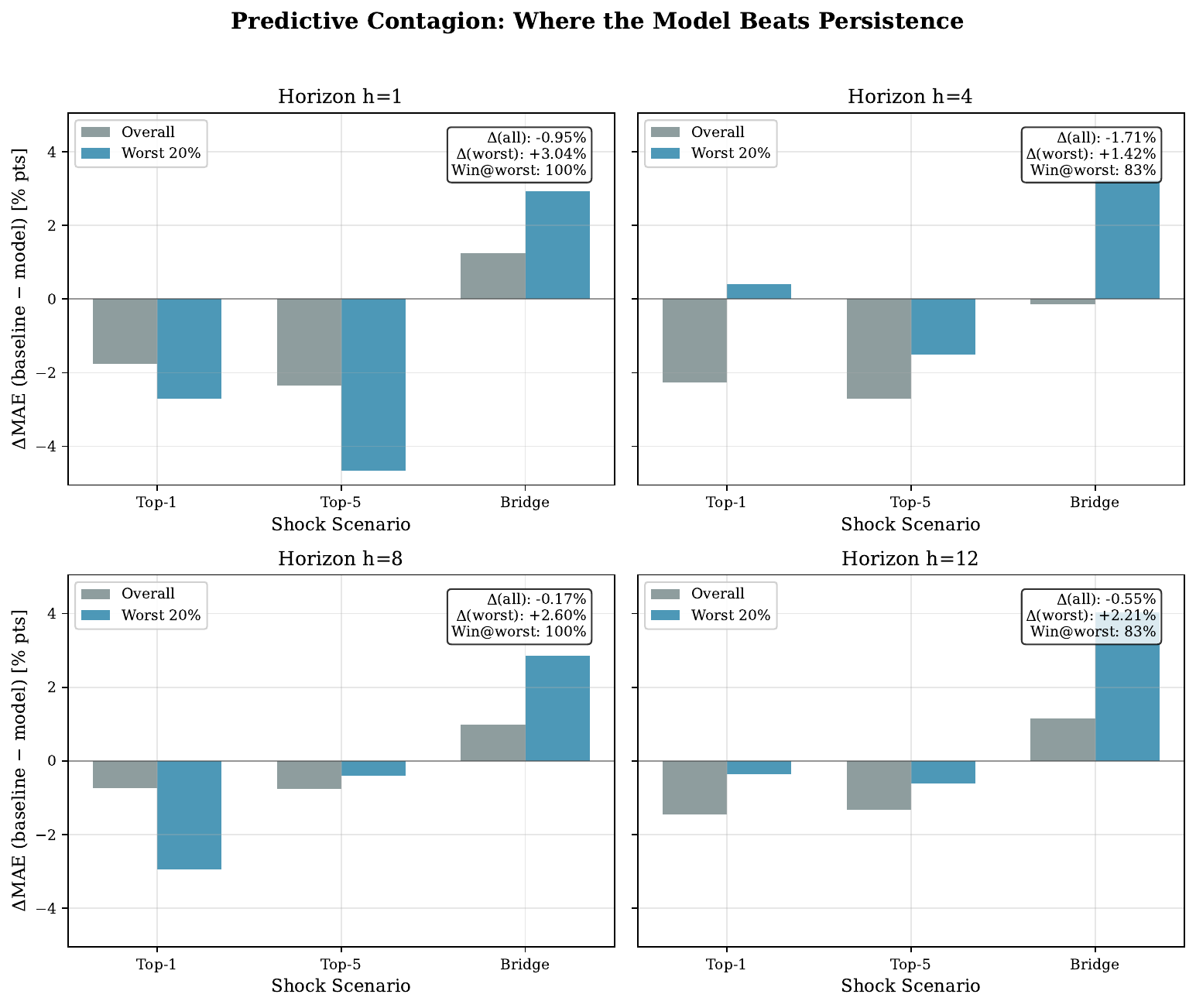}
\caption{Predictive contagion stress testing (Task II). Bars report $\Delta\mathrm{MAE}=\mathrm{MAE}(\text{persistence})-\mathrm{MAE}(\text{model})$ on system-loss percentage points. ``Overall'' is the full 2025 hold-out; ``Worst 20\%'' restricts to the 20\% largest persistence-baseline errors per horizon, pooled over shocks. Positive values indicate DeXposure-FM outperforms persistence; gains concentrate in the tail and under bridge-sector shocks where network structure shifts most.}
\label{fig:contagion-advantage}
\end{figure}

Figure~\ref{fig:contagion-comparison} shows the three-way comparison (baseline vs.\ model vs.\ realized) and Figure~\ref{fig:contagion-advantage} visualizes $\Delta\mathrm{MAE}$ in both the overall population and the worst 20\% tail. Table~\ref{tab:contagion-advantage} reports horizon-level summary statistics.

On the full hold-out, persistence is hard to beat ($\Delta\mathrm{MAE}<0$ on average), reflecting week-to-week stability: with 98.5\% mean edge overlap, the carry-forward graph is itself a near-perfect input to the simulator on the largest exposure links that drive contagion, so any imperfect forecast is at a structural disadvantage at the system-aggregate level. The value of a learned forecaster therefore emerges where this carry-forward assumption breaks down. In the worst-20\% tail, DeXposure-FM achieves consistently positive $\Delta\mathrm{MAE}$ with win rates of 83--100\%; gains are most pronounced under bridge-sector shocks, consistent with cross-chain contagion being sensitive to evolving network topology. The macroprudential reading is therefore selective: learned forecasts are most useful as a backstop for scenario analysis when exposures reallocate non-stationarily, i.e., in high-error regimes where a carry-forward graph is least reliable.

\begin{table}[t]
\centering
\small
\setlength{\tabcolsep}{6pt}
\begin{tabular}{lccc}
\hline
Horizon $h$ & $\Delta\mathrm{MAE}(\text{all})$ & $\Delta\mathrm{MAE}(\text{worst20\%})$ & Win rate (worst20\%) \\
\hline
1  & $-0.95$ & $+3.04$ & 100\% \\
4  & $-1.71$ & $+1.42$ & 83\% \\
8  & $-0.17$ & $+2.60$ & 100\% \\
12 & $-0.55$ & $+2.21$ & 83\% \\
\hline
\end{tabular}
\caption{Predictive stress testing summary (2025 hold-out). $\Delta\mathrm{MAE}$ is measured in system-loss percentage points. Worst20\% selects the 20\% largest persistence-baseline errors within each horizon (pooled across shock scenarios). ``Win rate'' is the fraction of worst20\% cases where the model has lower absolute error than persistence.}
\label{tab:contagion-advantage}
\end{table}

\section{Financial Economics Tools and Experimental Verifications}
\label{sec:econ}

%Section~\ref{sec:evaluation} establishes DeXposure-FM's predictive performance on machine learning benchmarks. 
Section~\ref{subsec:stability} evaluates contagion losses as a \emph{forecasting metric}: forecast graphs are scored against realized graphs through a fixed contagion simulator. This section instead uses the same forecast graphs as \emph{inputs to a measurement layer}: predicted graphs are translated into protocol watchlists, sectoral risk channels, and scenario losses that regulators or risk managers can inspect before the realized network is observed. %that connect the model's forecasts to macroprudential use cases.
%\subsection{Macroprudential Applications}
%\label{subsec:macro-applications}
The financial-economics contribution of DeXposure-FM is therefore not only better edge prediction, but a forward-looking macroprudential measurement layer for digital financial infrastructure.

Concretely, let $G_t$ denote the observed weekly exposure graph and let $\hat{G}_{t+h}$ denote the model's $h$-step-ahead forecast. We treat each monitoring ``tool'' as a deterministic functional of a graph (a measurement map) that can be applied either {descriptively} on $G_t$ or {predictively} via the \emph{forecast-then-measure} pipeline:
\begin{equation}
G_t \;\xrightarrow{\;\;\text{DeXposure-FM}\;\;}\; \hat{G}_{t+h}
\;\xrightarrow{\;\;\text{tools}\;\;}\;
\widehat{\mathcal{M}}_{t+h} \equiv \mathcal{M}(\hat{G}_{t+h}),
\end{equation}
where $\mathcal{M}(\cdot)$ includes protocol-level systemic-importance rankings (SIS), cross-sector spillover concentration, and scenario-based stress-test losses. This framing aligns with the macroprudential objective of producing \emph{actionable, time-$t$ signals} about \emph{time-$t+h$ fragility} (e.g., watchlists, risk channels, and scenario backstops), while remaining fully transparent about the underlying network objects. Its value is deliberately interpreted as \emph{conditional}: when the exposure network is stable, persistence is a strong monitoring baseline; when exposures reallocate, especially in tail-error weeks and bridge-sector shocks, the learned forecast provides the additional signal needed for macroprudential backstops.

\subsection{Systemic Risk Measurement: SIS and Spillovers}
\label{subsubsec:systemic-risk}

%\subsubsection{Economic object} 

%SIS produces a protocol-level systemic-importance ranking that combines interconnectedness and exposure concentration with protocol scale, while spillover monitoring summarizes cross-sector risk channels and their concentration.

We first introduce SIS and spillovers as systemic risk measurement.

\subsubsection{Implementation} 

We compute protocol-level and network-level indicators on each weekly exposure snapshot. For protocol $p$, the SIS combines interconnectedness, exposure concentration, and scale:
\begin{equation}
\text{SIS}_p = \alpha \cdot \widetilde{\text{PageRank}}_p + \beta \cdot \text{TailExposure}_p + \gamma \cdot \widetilde{\log(1+\text{TVL}_p)},
\label{eq:sis}
\end{equation}
where $\text{TailExposure}_p$ is the share of $p$'s top-$k$ outgoing exposures (we use $k=5$), and $\alpha,\beta,\gamma$ are non-negative weights that sum to 1 (default $\alpha=\beta=\gamma=1/3$). We normalize PageRank \citep{BrinPage1998PageRank} and $\log(1+\text{TVL})$ to comparable scales (denoted by tildes). In parallel, we aggregate edge weights into a sector-to-sector spillover matrix $\mathbf{S}\in\mathbb{R}^{K\times K}$ where $S_{ij}$ is total exposure from sector $i$ to $j$, and compute a scale-invariant spillover concentration index as the Herfindahl-Hirschman Index (HHI) \citep{Rhoades1993HHI} of off-diagonal entries:
\begin{equation}
\text{SpilloverIndex} = \text{HHI}\left(\{S_{ij} : i \neq j\}\right).
\end{equation}
Among the forecast-based indicators, network density and mean SIS are well-calibrated against realized values and support quantitative comparisons; HHI-based concentration is informative for directional changes rather than absolute levels, as shown by the level-vs-direction evidence in Figure~\ref{fig:forward-risk-scatter} and the event-window study in Figure~\ref{fig:early-warning}.

\begin{figure}[t]
\centering
\includegraphics[width=0.95\textwidth]{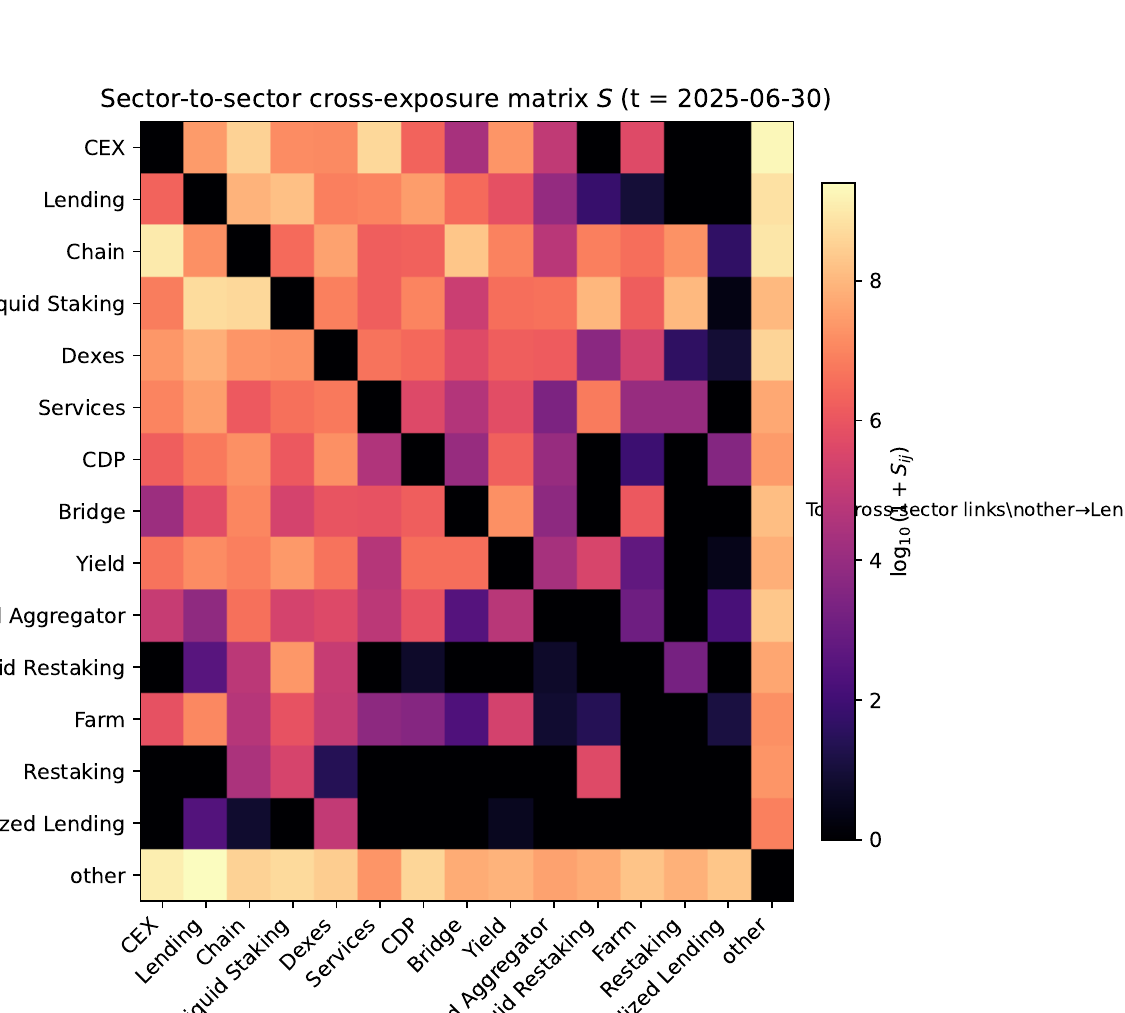}
\caption{Example visualization of the sector-to-sector cross-exposure matrix $\mathbf{S}$ computed on an observed snapshot $G_t$ (here $t=\text{2025-06-30}$). Cell colors show $\log_{10}(1+S_{ij})$, where $S_{ij}$ aggregates total exposure from sector $i$ to sector $j$.}
\label{fig:spillover-matrix}
\end{figure}

%\subsubsection{Use cases} 

These measures operationalize the network-theoretic insight that interconnectedness, not just size, determines systemic fragility \citep{Acemoglu2015SystemicRisk,GlassermanYoung2016Contagion}. Weekly SIS rankings help identify protocols whose distress would propagate broadly, while spillover monitoring can highlight sector pairs that act as risk conduits (e.g., stablecoins and bridges) \citep{KosseEtAl2023Stablecoin,ESRB2025,DieboldYilmaz2012}.

%\noindent\textbf{Why this is a model-enabled tool (not just a descriptive statistic).} 
SIS and spillovers are standard {descriptive} measures when computed on the observed network $G_t$. The contribution of DeXposure-FM is to make them \emph{forward-looking}: we first forecast a future exposure graph $\hat{G}_{t+h}$ and then compute \[\widehat{\text{SIS}}_{t+h}(p)=\text{SIS}_p(\hat{G}_{t+h}),\] \[\widehat{\text{SpilloverIndex}}_{t+h}=\text{SpilloverIndex}(\hat{G}_{t+h}).\] 
This yields actionable forecasts such as (1) a predicted watchlist of systemically important protocols at $t+h$, and (2) predicted cross-sector risk channels. 

At each week $t$, a supervisor computes the current dashboard on $G_t$ and the forecast dashboard on $\hat{G}_{t+h}$. Large predicted changes in the top-$K$ SIS list or spikes in predicted spillover concentration motivate targeted stress scenarios (Section~\ref{subsubsec:contagion}) and deeper qualitative review (e.g., governance risk and code audit status of newly elevated protocols).
\subsubsection{Empirical Validation}

Empirically, we validate this ``forecast-then-measure'' workflow at the level of aggregate risk metrics (including mean SIS and spillover concentration) by comparing values computed on $\hat{G}_{t+h}$ vs.\ $G_{t+h}$ (Figure~\ref{fig:forward-risk-scatter}) and by event-window case studies (Figure~\ref{fig:early-warning}).

%\noindent\textbf{Open-source tool.} We release an implementation of this workflow in \texttt{run\_macroprudential\_tools.py} (CLI) and \texttt{src/macroprudential\_tools.py} (library). The CLI supports both \texttt{observed} mode (compute tools on $G_t$) and \texttt{predict} mode (forecast $\hat{G}_{t+h}$ and then compute the same tools on predicted and realized graphs when available), and outputs JSON for easy integration into monitoring pipelines.

\subsection{Stress Testing and Contagion Assessment}
\label{subsubsec:contagion}

%\noindent\textbf{Economic object.} 
Stress testing maps a counterfactual shock scenario into a system-wide loss measure by propagating distress through the exposure network, providing a scenario-specific analogue of macro stress tests in a DeFi setting.

\subsubsection{Implementation} 

We implement a DebtRank-style contagion simulation to assess systemic loss propagation, following the loss-allocation logic \cite{EisenbergNoe2001,BattistonEtAl2012DebtRank}. Given an initial shock to protocol $p_0$ with loss ratio $\delta_0$, we run the following procedure:
\begin{enumerate}
  \item \emph{Initialize:} $\text{Loss}_{p_0} = \delta_0 \cdot \text{TVL}_{p_0}$.
  \item \emph{Propagate:} When a distressed debtor $p$ has loss $\text{Loss}_p$, we allocate its loss to its creditors $q$ proportionally to their exposures $E_{qp}$:
  \begin{equation}
    \Delta \text{Loss}_q \;=\; \text{Loss}_p \cdot \frac{E_{qp}}{\sum_{q'} E_{q'p}}.
  \end{equation}
  \item \emph{Iterate:} If $\text{Loss}_q > \tau \cdot \text{TVL}_q$ (distress threshold $\tau=0.1$), protocol $q$ becomes distressed and propagates losses to its creditors. Losses are capped at $\text{TVL}_q$.
  \item \emph{Terminate:} When no new protocols become distressed.
\end{enumerate}
We report total system loss (as a percentage of total TVL), contagion depth (propagation rounds), and affected-protocol counts. The three stress scenarios used throughout are: top protocol shock (50\% TVL loss), top-5 protocols shock (30\%), and bridge sector shock (100\%).

\subsubsection{Empirical Verification}

Section~\ref{subsec:stability} evaluates \emph{forward-looking} stress testing; here, we run the same simulator on $G_t$ (persistence baseline), $\hat{G}_{t+h}$ (model forecast), and $G_{t+h}$ (realized future), and then compare the resulting system loss. Because the DeXposure network is highly persistent week-to-week, the persistence baseline is strong \emph{on average}. However, in the subset of weeks where realized losses deviate most from persistence (the worst 20\% baseline-error regime), DeXposure-FM provides consistent improvements (Table~\ref{tab:contagion-advantage}), with gains concentrated in bridge-sector shocks (Figure~\ref{fig:contagion-advantage}). This supports a macroprudential interpretation: the model is most valuable as a \emph{backstop} for scenario analysis under non-stationary reallocation of exposures.

\subsection{Forward-looking Risk Metric Forecasting}
\label{subsubsec:forward-looking-risk}

%\noindent\textbf{Economic object.} 
Aggregate risk metrics (e.g., concentration, density, spillovers) serve as dashboard-style indicators of evolving fragility. Forecasting these aggregates provides a practical way to monitor trend shifts and calibrate the forecast-then-measure pipeline.

\subsubsection{Implementation} 

Beyond stress testing, we can forecast aggregate stability indicators by first predicting a future exposure graph $\hat{G}_{t+h}$ and then computing deterministic risk metrics on it (e.g., TVL concentration, edge concentration, network density, spillover index, and summary statistics of SIS). We compare these predicted metrics to their realized counterparts on $G_{t+h}$ across the 2025 hold-out set.

\subsubsection{Empirical Verification}

\begin{figure}[t]
\centering
\includegraphics[width=0.95\textwidth]{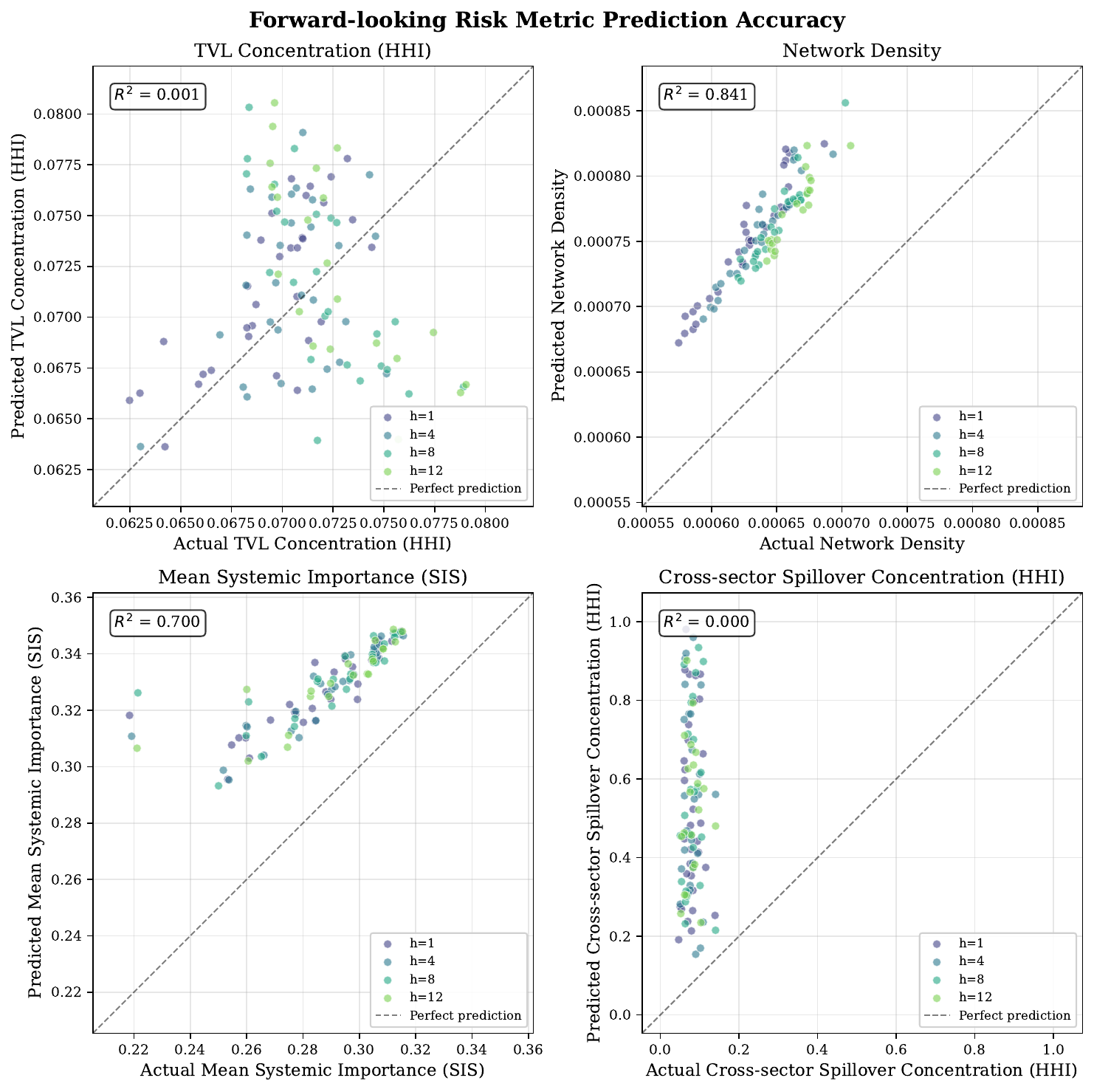}
\caption{Forward-looking risk metric forecasting on predicted exposure graphs. Each point compares a realized metric on $G_{t+h}$ (x-axis) to the same metric computed on the predicted graph $\hat{G}_{t+h}$ (y-axis), across horizons.}
\label{fig:forward-risk-scatter}
\end{figure}

Figure~\ref{fig:forward-risk-scatter} reports predicted vs.\ realized risk metrics for $h\in\{1,4,8,12\}$. We interpret these plots as a \emph{calibration and monitoring} check: even when persistence dominates levels, the predicted metrics can flag directional changes and large deviations that motivate deeper scenario analysis and qualitative review.

\subsection{Early Warning Signals and Event Studies}
\label{subsubsec:shock-events}

%\noindent\textbf{Economic object.} 
Early warning analysis evaluates whether forecast-based measurements shift in advance of major stress events, providing event-study evidence that the forecast-then-measure tools can produce timely alerts around regime changes.

\subsubsection{Implementation} 

We examine two major stress events: the Terra/Luna collapse (May 9, 2022) and the FTX collapse (November 7, 2022). To avoid event leakage, for each event, we train an event-specific forecaster using only snapshots strictly prior to the event window, then generate one-step-ahead forecasts within the event window. We track structural indicators and stress-test outcomes (via Section~\ref{subsubsec:contagion}) and monitor potential early warning features:
\begin{itemize}
  \item Changes in network connectivity (e.g., density $\rho_t$),
  \item Concentration shifts (e.g., TVL HHI), with practical threshold rules such as $\Delta\text{HHI}_t > 2\sigma$,
  \item Predicted stress-test losses under standard scenarios (Section~\ref{subsubsec:contagion}).
\end{itemize}

\subsubsection{Empirical Verification}

\begin{figure}[t]
\centering
\includegraphics[width=0.95\textwidth]{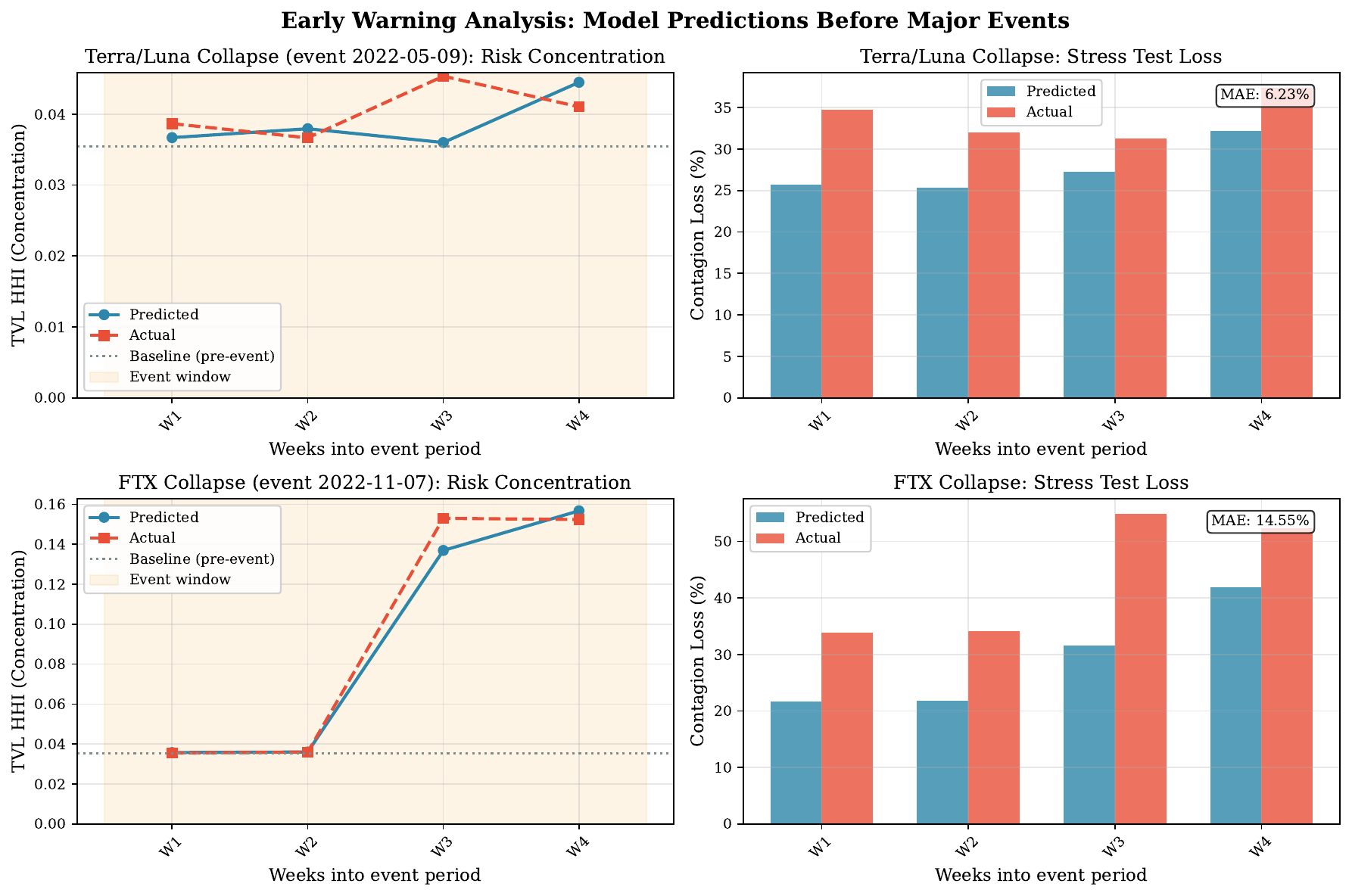}
\caption{Early warning event studies. For each event window, we compare predicted vs.\ realized risk concentration (HHI) and predictive stress-test loss (contagion system loss \%).}
\label{fig:early-warning}
\end{figure}

Table~\ref{tab:shock-results} summarizes structural shifts during the event windows, and Figure~\ref{fig:early-warning} compares predicted vs.\ realized trajectories for risk concentration and stress-test loss. These case studies illustrate how model-based forecasting can complement descriptive monitoring by providing forward-looking alerts around regime changes.

\begin{table}[t]
\centering
\small
\begin{tabular}{lcccc}
\hline
\textbf{Event} & \textbf{Pre-TVL (\$B)} & \textbf{Post-TVL (\$B)} & \textbf{$\Delta$TVL} & \textbf{$\Delta$Edges} \\
\hline
Terra/Luna & 261.9 & 166.3 & $-$36.5\% & +9.3\% \\
FTX & 119.9 & 175.1 & +46.1\% & $-$1.8\% \\
\hline
\end{tabular}
\caption{Network structure changes during stress events, comparing the last pre-event week (anchor) to the final week of the event window. Terra/Luna shows a flight-to-quality pattern with TVL collapse but edge count increase as capital reorganizes across protocols. The FTX event shows a \emph{positive} $\Delta$TVL because our dataset captures only on-chain DeFi protocols: the collapse of a centralized exchange triggered capital migration \emph{into} DeFi as users withdrew funds to self-custody, consistent with the ``flight to decentralization'' narrative \citep{VidalTomasEtAl2023FTX}.}
\label{tab:shock-results}
\end{table}

\subsection{Validity and Scope}
\label{subsec:validity}

Model-based risk measures are only as reliable as the conditions under which they were trained and validated. We report several boundaries that users should keep in mind.

\subsubsection{Stable vs. Turbulent Regimes}
The DeXposure dataset exhibits 98.5\% mean edge overlap week-to-week, so persistence is a strong baseline for many magnitude targets. During structural breaks, such as new protocol launches, governance attacks, or rapid deleveraging, the historical edge distribution may shift abruptly, and forecasts should be interpreted with greater caution. The stress-testing results in Table~\ref{tab:contagion-advantage} suggest that learned forecasts are most useful in high-error tail regimes where persistence is least reliable.

\subsubsection{Data Coverage}
DeXposure-FM is trained exclusively on on-chain DeFi protocols sourced from DeFiLlama. It does not observe:
\begin{itemize}
  \item Centralized exchange (CEX) balances or order-book exposures,
  \item Over-the-counter (OTC) derivatives or lending agreements,
  \item Off-chain reserves backing stablecoins (e.g., Treasury holdings of USDC).
\end{itemize}
Consequently, the model captures intra-DeFi contagion but cannot directly assess spillovers to or from traditional financial institutions, a gap that matters as DeFi-TradFi linkages deepen.

\subsubsection{Temporal Resolution}
The training dataset aggregates to weekly snapshots. Intraday or even daily dynamics, such as flash-loan attacks \citep{Qin2021FlashLoans} or rapid liquidation cascades \citep{Warmuz2023ToxicLiquidationSpirals}, are smoothed out. For high-frequency risk monitoring, complementary tools operating on block-level data would be required.

\subsubsection{Cross-chain Data Consistency}
DeXposure aggregates data across 602 blockchains, each with its own indexing quirks and oracle delays. Minor inconsistencies in timestamp alignment or token-price feeds can introduce noise. Users should be aware that edge weights near reporting boundaries may reflect data artifacts rather than genuine exposure changes.

\subsubsection{TVL as a Measurement Proxy}
TVL aggregates heterogeneous assets without risk weights, a limitation we discuss in Section~\ref{sec:conclusion}. Nevertheless, TVL remains the main metric for assessing DeFi protocol significance \citep{BIS2025TVL}, and its changes over time capture the capital flows that drive contagion \citep{ZhangEtAl2026TVLDynamics}. The tools are therefore most informative for relative monitoring (rankings, concentration changes, scenario comparisons) rather than absolute loss accounting.

\subsubsection{Our aim: Complement, not replace, qualitative judgment}
Model outputs such as SIS rankings and spillover indices are quantitative summaries, not definitive verdicts. A protocol may rank low on SIS yet pose idiosyncratic risks (e.g., smart-contract vulnerabilities, oracle manipulation) that network topology alone cannot capture \citep{DuleyEtAl2023Oracle,gogol2024sokdecentralizedfinancedefi}. Supervisors should triangulate model signals with code audits, governance analyses, and market intelligence.

\section{Future  Development Plan}

We plan to continuously update and improve DeXposure-FM. Below is our future development plan.

%\begin{itemize}
    %\item 
    
\subsection{Enlarging Training Data Pool} 
The current DeXposure-FM is trained on weekly inter-protocol snapshots from the DeXposure dataset, using a GraphPFN encoder as pretrained initialization. While this provides broad on-chain coverage at a stable temporal resolution, it omits key off-chain channels through which DeFi is funded, hedged, and linked to traditional markets. In DeXposure-FM~2.0, we plan to expand the training pool by incorporating additional data sources, including centralized exchange (CEX) balances and order-book–implied exposures, over-the-counter (OTC) derivatives and lending agreements, and disclosures on off-chain stablecoin reserves (e.g., USDC’s Treasury-backed holdings). We also aim to add higher-frequency data (daily or intraday, where feasible), particularly around stress episodes, to better capture rapid deleveraging and liquidity dynamics that weekly aggregation smooths out.

\subsection{Expanding Proxies for Credit Exposure}
Total Value Locked (TVL) is a convenient but imperfect proxy for economic exposure. It aggregates heterogeneous assets into a single dollar value without adjusting for liquidity, haircuts, or collateral quality. As a result, equal TVL can imply very different loss-absorbing capacity and contagion risk; for example, \$1B held in thinly traded governance tokens is not economically equivalent to \$1B in USDC, and TVL can also move mechanically with prices even when underlying positions are unchanged. To address these limitations, DeXposure-FM~2.0 will incorporate asset-level risk weights and composition features (e.g., liquidity and volatility proxies, depeg risk, and observable collateral rules) to produce risk-adjusted exposure estimates and more interpretable systemic-risk indicators.

\subsection{Model Drift and Continuous Update}
DeFi evolves rapidly: new chains launch, token standards change, and incentive mechanisms shift. This non-stationarity implies that a model trained on historical regimes may gradually lose accuracy as exposures and flows depart from past patterns, including both gradual structural change and discrete breaks from major upgrades or new collateral rules. Periodic retraining on updated snapshots is therefore essential to maintain forecast quality, ideally paired with drift monitoring (e.g., calibration and error shifts by sector/chain) and versioned releases so that risk signals can be traced to specific data vintages and model iterations.

\subsection{Innovative Model Architecture}
The future DeXposure-FM 2.0 will include an innovative hybrid architecture that combines the current direct multi-horizon forecasting paradigm \citep{RasulEtAl2023LagLlama} with diffusion-based generative modeling \citep{MeijerChen2024DiffusionTS}. Auto-regressive models are efficient and accurate for point and conditional forecasts, but they can struggle to represent globally coherent network trajectories over long horizons, especially under stress. Diffusion models, by contrast, provide a principled way to learn complex conditional distributions by gradually denoising samples, and have recently shown strong performance in structured generation tasks, such as in diffusion language models \citep{LiEtAl2022DiffusionLM}. %In DeXposure-FM 2.0, we plan to use an autoregressive backbone to produce fast, state-dependent predictions of key sufficient statistics (e.g., protocol states and coarse exposure aggregates), while a conditional diffusion module refines these into full network realizations—jointly sampling link existence and weights in a way that preserves structural constraints such as sparsity, sectoral organization, and budget/flow consistency. This hybrid design would allow the model to output calibrated predictive distributions (not just point forecasts), support scenario-conditional stress testing via controlled guidance, and improve robustness to regime shifts by explicitly modeling the space of plausible future network configurations rather than committing early to a single trajectory.

\subsection{Open Source, Competitions, and Community Development}
We endeavor to develop DeXposure-FM to become fundamental infrastructure rather than a closed research artifact. We have released an open-source version of model weights and training code. We will continue our open-source efforts when improving the model. Based on the DeXposure dataset and DeXposure-FM model, we will organize public benchmarking competitions. These challenges will be designed to encourage robust methods and will include baseline implementations, leaderboards, and model cards reporting calibration and failure modes. We also aim to build a sustained community around DeFi economic measurement. By lowering barriers to entry and making the measurement process transparent, we hope to accelerate cumulative progress on reliable, policy-relevant tools for monitoring systemic risk in decentralized financial networks.

\section{Conclusions}
\label{sec:conclusion}

DeXposure-FM frames DeFi credit exposures as a domain-specific GFM problem for dynamic graph-tabular financial networks. Initializing a GraphPFN-based encoder on weekly exposure graphs and protocol descriptors yields strong gains for edge-existence forecasting and improves edge-weight prediction over learned neural baselines, while persistence remains a demanding comparator for highly stable magnitude targets. The forecast-then-measure pipeline restores the financial-economics objective: predicted graphs become forward-looking measurements of systemic importance, sector spillovers, and stress-test losses, especially in tail regimes where the carry-forward assumption is least reliable. Future work includes zero- and few-shot transfer to other financial graphs such as interbank exposures, supply-chain credit, and cross-CEX flows; risk-weighted exposure proxies beyond TVL; higher-frequency data integration; drift-aware continuous retraining; and distributional graph forecasting for scenario analysis.

\bibliographystyle{elsarticle-harv} 
\bibliography{sections/DeXposure-FM}



%\begin{thebibliography}{00}

%% For numbered reference style

%\bibitem{lamport94}
%  \textit{\LaTeX: a document preparation %system},
%  Addison Wesley, Massachusetts,

%\end{thebibliography}

\end{document}